\documentclass[11pt]{article}

\usepackage[final]{acl}

\usepackage{times}
\usepackage{latexsym}
\usepackage{multirow}
\usepackage[T1]{fontenc}
\usepackage[utf8]{inputenc}

\usepackage{microtype}

\usepackage{inconsolata}

\usepackage{graphicx}

\title{Nuha-Speech: Building General-Purpose Arabic Speech-LLMs}

\author{Yingzhi Wang \\
  Elm Company, KSA \\
  \texttt{ywang@elm.sa} \\\And
  Reem Alhazzani \\
  Elm Company, KSA \\
  \texttt{ralhazzani@elm.sa} \\\And
  Muhammad Alqurishi \\
  Elm Company, KSA \\
  \texttt{mualqurishi@elm.sa} \\}

\begin{document}
\maketitle
\begin{abstract}
As Speech Large Language Models (speech-LLMs) become increasingly multilingual, Arabic remains significantly underrepresented, highlighting the need for dedicated infrastructure to train and evaluate Arabic speech-LLMs.

To address this gap, we introduce Nuha-Speech, a comprehensive initiative to develop general-purpose Arabic speech-LLMs spanning dataset construction, model training, and systematic evaluation. Specifically, we constructed a large-scale Arabic Speech Question-Answering (SQA) corpus comprising over 1.5 million training samples to allow instruction tuning over a broad range of core speech tasks. Then, the corpus was used for supervised fine-tuning based on Qwen-Omni model variants at different scales. Finally, we designed an evaluation framework featuring diverse tasks and tailored metrics. Through this work, we aim to establish foundational infrastructures for Arabic Speech-LLMs under constraints imposed by limited Arabic speech resources.
\end{abstract}

\section{Introduction}

Speech Large Language Models (Speech-LLMs) have seen swift advancements in recent years. However, so far, few models provide support for Arabic. The Octopus family models \cite{octopus} introduce Arabic-centric Speech-LLMs that can handle three tasks: Automatic Speech Recognition(ASR), Arabic–to-English speech translation, and dialect identification. The models follow a Salmonn-style \cite{salmonn} architecture incorporating both semantic and acoustic encoders, and include a distilled variant that achieves competitive performance. While Octopus represents a valuable effort toward enhancing Speech-LLMs with Arabic capabilities, compared to recent Speech-LLM advances, its task coverage remains very limited and lacks zero-shot generalization, limiting its utility as a general-purpose Speech-LLM. Furthermore, its reliance on predominantly private datasets restricts its reproducibility and broader community adoption.

% The three Qwen models all accept audio input and are instruction-tuned, but they differ in multilingual scope and Arabic support. 
The Qwen family has released a series of high-performance speech-LLMs, with the level of Arabic language support varying from model to model.
Qwen2-Audio\cite{qwen2_audio} is a popular speech-LLM designed for voice-chat and audio analysis. It outperforms prior models on instruction-following benchmarks. Although Qwen2-Audio relies on Whisper-large-v3\cite{whisper}, which supports Arabic among many other languages, all available demos and test cases are only in English and Chinese, there is no explicit proof that the model can follow instructions given Arabic speech inputs. Qwen2.5-Omni\cite{qwen2.5_omni} similarly demonstrates strong end-to-end speech instruction following in over 29 languages, benefiting from its diverse multilingual training data. However, its public evaluation and examples have focused on English and Chinese, its ability in Arabic remains unverified and must be inferred. By contrast, Qwen3-Omni\cite{qwen3_omni} stands out as explicitly multilingual, including Arabic among the 19 supported speech-input languages. This enables strong Arabic speech comprehension across a wider range of linguistic contexts than most comparable models.

% In effect, Qwen3-Omni is natively built to understand Arabic speech. By contrast, the earlier Qwen models depend on Whisper’s broad language support and do not explicitly show Arabic instruction-following capability.

The scarcity of Arabic-capable speech-LLMs reflects several deep challenges. First, when considering non-ASR speech tasks, the number of publicly accessible Arabic speech corpora remains extremely limited. For instance, to date, there remains a striking lack of public Arabic speech corpora of sufficient scale for tasks such as Speech Emotion Recognition (SER) or Speech Question Answering (SQA). In addition, Arabic speech instruction-tuning data is extremely scarce: the majority of speech instruction-following corpora are in English, and although a few have been translated into Arabic, genuine Arabic speech–instruction pairs are almost non-existent. Second, there is no widely adopted evaluation benchmark for Arabic speech LLMs. The existing benchmarks remain heavily centered on mainstream languages like English \cite{dynamic_superb, air_bench, audiobench, mmsu}, with most speech tasks derived from English datasets, making it difficult to assess models' understanding of less-represented languages like Arabic.

To fill this gap, we propose Nuha-Speech, a comprehensive effort aimed at building general-purpose Arabic speech LLMs via dataset construction, model training, and benchmark development. 
By combining public and curated datasets, we assembled a training dataset of 1.5 million Arabic speech instruction-following samples. This dataset spans core speech tasks covering both speech understanding and speech paralinguistics. Moreover, we leveraged mostly publicly available datasets and transparent, reproducible curation strategies, facilitating dataset replication by the research community.
% We also release a toy set to provide examples of the dataset structure and prompting.
We then used this dataset to fine-tune three Qwen-Omni model variants at different parameter scales, aiming to equip them with full Arabic comprehension capabilities. Finally, we constructed test sets for all involved tasks, adopted different evaluation metrics to form a multi-task benchmark, and evaluated models in both pre- and post-fine-tuning settings. The results demonstrate that the fine-tuned models achieved clear improvements across all listed tasks.

\section{Nuha-Speech Dataset}

In this section, we outline how the training corpus was constructed for each task. Table 1 provides an overview of the collected datasets, including tasks, sources, and sizes. To unify all tasks within a shared semantic space, each sample is formatted as a \{speech, instruction, output\} tuple. In this work, we exclusively focus on training models with text-only outputs. For each task, we also designed a broad instruction set using GPT-5 to better support zero-shot generalization and avoid overfitting to narrow prompts. 
% We adopt both 3B and 7B models of Qwen2.5-omni to enrich our candidate pool and facilitate the comparison of performance across different model scales.

\begin{table}[htbp]
  \caption{Overview of the training corpus for Nuha-Speech.}
  \label{tab:environments}
  \centering
  \resizebox{\columnwidth}{!}{
    \begin{tabular}{c c c}
    \hline
    \textbf{Task} & \textbf{Data Source} & \textbf{\#Samples}\\
    \hline
    \multirow{4}{*}{ASR} & MGB-2 & 278K \\
     & MASC & 263K \\
     & SADA & 171K \\
     & Common Voice & 78K \\
    \hline
    \multirow{2}{*}{AST} & CoVoST-v2 (Ar2En) & 5K \\
     & ASR-Curated & 150K \\
    \hline
    \multirow{2}{*}{SQA} & MGB-2 Closed-ended & 150K \\
     & MGB-2 Open-ended & 150K \\
    \hline
    DI & ADI-17 & 150K \\
    \hline
    SER & Elevenlabs-Syn & 20K \\
    \hline
    \multirow{3}{*}{AR} & SADA & 5K \\
     & Common Voice & 5K \\
     & Elevenlabs-Syn & 10K \\
    \hline
    GR & SADA & 51K \\
    \hline
    \multirow{2}{*}{Multi-SQA} & SADA & 5K \\
     & Emotion Reasoning & 10K \\
    \hline
    Total & & 1501K \\
    \hline
    \end{tabular}
}
\end{table}

\subsection{Automatic Speech Recognition (ASR)}
ASR serves as one of the most fundamental speech tasks. In our training, ASR is also regarded as the primary objective. We selected four large-scale, multi-dialect public datasets (SADA\cite{sada}, Common Voice\cite{commonvoice}, MASC\cite{masc}, and MGB-2\cite{mgb2}) and removed segments outside the 1–10s range to optimize training efficiency. This yields roughly 790K ASR training samples in total.

% For Arabic ASR evaluation, we adopt the existing Open Universal Arabic ASR Leaderboard \cite{}, a benchmark framework spanning multiple Arabic dialects across 6 different datasets. It consists of 46757 test samples in total. The average word error rate (WER) and Character Error Rate (CER) across all the test sets are employed as the reported metrics. 

\subsection{Automatic Speech Translation (AST)}
Automatic Speech Translation (AST) maps source-language spoken input to target-language textual translation in an end-to-end way. 
% It demands both strong comprehension of the speech and cross-language semantic alignment. 
AST is highly valuable in real-world workflows for Arabic, especially when translating into English.
In our work, we adopt the Arabic-to-English split of the widely used CoVoST-v2 \cite{covost_v2} dataset.
% which consists of 2334 test samples. 
Due to its very limited number of samples, we then randomly selected 150K samples from the ASR dataset collected above and translated their transcriptions into English using the Qwen3-32B\footnote{\url{https://huggingface.co/Qwen/Qwen3-32B}} model to obtain a sufficient number of training samples.
% We calculate both BLEU and sentence similarity as the evaluation metrics. BLEU captures surface overlap and alignment with reference translations, while sentence similarity helps assess deeper semantic equivalence beyond exact matches. For computing sentence similarity, we use the Gemma 3 \cite{} embedding model of 300M parameters \footnote{}, which is trained on over 100 languages and particularly designed for high-quality multilingual semantic embeddings.

\subsection{Speech Question Answering (SQA)}
% The SQA task requires generating answers to questions based on a spoken context. 
In speech understanding, the SQA task focuses on querying the semantic and factual information given spoken content. It closely aligns with how humans naturally interact with conversational systems. However, there is currently no suitable Arabic dataset that includes both spoken context and textual QA pairs. Given the rich content of the MGB-2 dataset, we utilized its transcriptions as input to generate 150K closed-ended and 150K open-ended QA pairs using the Qwen3-32B model. The closed-ended QA focuses on enhancing structured query handling and exact content retrieval, while the open-ended QA promotes higher-level skills such as cross-sentence reasoning and expressive language generation.

% Consequently, we base on the SD-QA \cite{} textual QA dataset and apply a state-of-the-art open-source TTS model to convert the text context into speech across 58 speakers voices. To ensure the TTS quality, we also used Whisper-Large-v3 \cite{} to transcribe the generated speech, filtered samples based on CER, and ended up with 1,237 high-quality curated samples.

% To evaluate the SQA task, we employ the model-as-judge approach with Llama-3.3-70B-Instruct \cite{}, rating the model’s response considering the question, context, and reference answer. 简单介绍一下评分prompt, 详细可见github.

\subsection{Dialect Identification (DI)}
Dialect Identification is a crucial task in Arabic speech processing, since one of Arabic’s defining features is its diverse dialects. Accurate DI is therefore important for building robust Arabic speech technologies, particularly in multilingual and multi-dialectal settings where dialectal variation can significantly affect downstream performance. For the DI task, we randomly selected 150K training samples from the well-established ADI-17 dataset, which is designed to classify 17 different Arabic dialects.

% For this task, well-established large datasets exist. We have incorporated the ADI-17 dataset \cite{}, of which the test split contains 12150 samples with 17 balanced dialect classes. When evaluating the accuracy, any response that falls outside the domain or predicts a non-included dialect is considered a misclassification.

\subsection{Speech Emotion Recognition (SER)}
Speech emotion recognition in Arabic has long been constrained by the lack of large-scale open-source datasets. In our work, using an emotion-conversion TTS model from Elevenlabs\footnote{\url{https://elevenlabs.io/blog/introducing-turbo-v25}}, we synthesized 20K emotional samples across four balanced categories: angry, happy, sad, and neutral. Specifically, each category's samples were randomly synthesized by 8 separate speakers in order to reduce the risk of speaker-specific bias. To ensure the effectiveness of emotion conversion, we activated style exaggeration and enforced stronger speaker similarity control during the synthesis process.
% Since no suitable open-source large-scale Arabic speech emotion dataset exists, we opted to synthesize emotional speech via a controllable TTS system \cite{}. We generated 4000 samples in four balanced categories: sad, happy, angry, and neutral, and measured accuracy to assess the performance.

\subsection{Age Recognition (AR)}
Similarly, few Arabic speech datasets contain age annotations, and existing ones exhibit a strong bias toward adult and middle-aged speakers, offering minimal coverage of younger and elderly populations. In our work, we defined age classification into three groups: young (under 18), adult (18–60), and elder (above 60). We extracted all available young and elder samples from SADA and Common Voice, then added a balanced number of adult samples, resulting in a dataset of 10K samples. Additionally, we synthesized a balanced set of speech samples for all three age groups using the same TTS model from ElevenLabs, with 10 distinct speakers per group, resulting in another 10K samples.

% For this task, well-established large datasets exist. We have incorporated the ADI-17 dataset \cite{}, of which the test split contains 12150 samples with 17 balanced dialect classes. When evaluating the accuracy, any response that falls outside the domain or predicts a non-included dialect is considered a misclassification.

% To mitigate this, in addition to using the existing Common Voice 18.0 dataset, we also synthesized speech for different age groups using the controllable TTS system \cite{}. In the end, we obtained 3000 samples in total for young, adult, and elderly speakers, and we report accuracy on this three-class classification task.

\subsection{Gender Recognition (GR)}

Given the observed cross-lingual variability in gender recognition performance\cite{gender_research1}, we also included the GR task into our model training. A total of 51K samples were extracted from the SADA dataset, with an equal distribution of male and female speakers.

\subsection{Multi-SQA}
To enhance the model's robustness and adaptability in complex spoken language scenarios, we also incorporated the multi-turn SQA task. First, leveraging the gender, age, and ASR annotations available in the SADA dataset, we constructed speech-analytics-style multi-turn SQA samples by merging these three isolated tasks, resulting in 5K samples with a balanced distribution across gender and age groups.

Second, a further multi-turn SQA task was designed that focused on emotion reasoning, where the model is first asked to recognize the speech emotion, then transcribe the spoken content, and finally infer the emotion correlation between the two modalities. We employed Qwen3-32B to generate 10K such cross-modality reasoning QA pairs based on designed SER dataset and its transcriptions.

\section{Instruction Tuning}
This section presents the baseline models and the instruction tuning setup.

\subsection{Baseline Models}
We used the Qwen-Omni model series as our fine-tuning baseline, taking advantage of the built-in speech input support and speech understanding capabilities, along with the Qwen LLM backbone's proven generalization strength for low-resource languages like Arabic\cite{cameleval, araeval}. The included models are: Qwen2.5-omni-3B\footnote{\url{https://huggingface.co/Qwen/Qwen2.5-Omni-3B}}, Qwen2.5-omni-7B\footnote{\url{https://huggingface.co/Qwen/Qwen2.5-Omni-7B}} and Qwen3-omni-30B\footnote{\url{https://huggingface.co/Qwen/Qwen3-Omni-30B-A3B-Instruct}}.

Qwen2.5-Omni adopts the Whisper-large-v3 encoder used in Qwen2-Audio and is powered by LLM from the Qwen2.5 series. Qwen3 improves upon this by integrating a more sophisticated AuT audio encoder and leveraging a more modern, efficient Mixture-of-Experts (MoE) architecture from the Qwen3 model series. We incorporated both the 3B and 7B variants of Qwen2.5 into our candidate model pool in order to broaden the range of evaluated models and to enable a more direct and systematic comparison of performance across different model scales.

\subsection{Experimental Setups}

\begin{table}[htbp]
  \caption{Nuha-Speech training curriculum.}
  \label{tab:environments}
  \centering
  \resizebox{\columnwidth}{!}{
    \begin{tabular}{c c c c c}
    \hline
    \textbf{Stage} & \textbf{Task} & \textbf{Data} & \textbf{Module} & \textbf{\#Epochs}\\
    \hline
    1 & ASR & ASR-790K & LLM (Lora) & 3 \\
    \hline
    % 2 & All & ASR-200K + Other-711k & LLM (Lora)\\
    % 2 & All & \multirow{2}{*}{\shortstack{C\\D}} & LLM (Lora) \\
    % &   &                                   &   \\
    \multirow{2}{*}{2} & \multirow{2}{*}{All} & \multirow{2}{*}{\shortstack{ASR-200K\\Others-711K}} & \multirow{2}{*}{LLM (Lora)} & \multirow{2}{*}{2} \\ & & & & \\
    \hline
    \end{tabular}
}
\end{table}

Following previous studies\cite{qwen2_audio, das2024speechverse, du2025making}, we adopted a two-stage training curriculum for Nuha-Speech, as illustrated in Table 2. In the first stage, the model was trained only to perform the ASR task, aiming to establish a strong audio-text mapping as the foundation for subsequent speech understanding tasks. In the second stage, we trained the models on all tasks. Given that the models had already acquired ASR capabilities in stage 1, we reduced the ASR training samples from 790K to a randomly sampled 200K subset. This not only helped preserve ASR performance but also reduced training time and improved balance across tasks. Recognizing that the audio encoders in all baseline models can already effectively process Arabic speech, we froze the encoders and trained only the LLM with LoRA\cite{lora} in both stages to stabilize training. According to the Arabic ASR leaderboard\footnote{\url{https://huggingface.co/spaces/elmresearchcenter/open\_universal\_arabic\_asr\_leaderboard}}, the Qwen3-Omni-30B model has demonstrated competitive ASR performance, ranking among the top-rated models in the benchmark. Therefore, for Qwen3-Omni-30B model we skipped the ASR-only Stage 1 and proceeded directly to Stage 2. In contrast, both Stage 1 and Stage 2 were applied to Qwen2.5-Omni-3B and Qwen2.5-Omni-7B models to complete the full training pipeline.

More specifically, we trained Qwen2.5-Omni-3B and Qwen2.5-Omni-7B using a global batch size of 16 and a learning rate of 1e-4 throughout both stages. For Qwen3-Omni-30B, we adopted a relatively smaller batch size of 8 while maintaining the same learning rate, and set the MoE router’s load balancing loss coefficient to 1e-3. LoRA was configured with a rank of 8, alpha of 32, and applied to all linear layers. Finally, corresponding to the parameter sizes of their respective baseline models, we refer to our fine-tuned models as Nuha-Speech-3B, Nuha-Speech-7B, and Nuha-Speech-30B.

\section{Nuha-Speech Benchmark}

In this section, we describe the construction of the test datasets and Nuha-Speech benchmark, followed by a detailed analysis of the experimental results.

\subsection{Test Sets and Metrics}
To comprehensively evaluate model performance, we adopted the same 7 tasks involved during the training stage, covering both speech understanding and speech paralinguistics. Table 3 summarizes the detailed evaluation setup, including tasks, datasets, metrics, and the number of samples.

\begin{table}[htbp]
  \caption{Overview of the evaluation setups: tasks, datasets, metrics, and number of samples.}
  \label{tab:environments}
  \centering
  \resizebox{\columnwidth}{!}{
    \begin{tabular}{c c c c}
    \hline
    \textbf{Task} & \textbf{Data Source} & \textbf{Metric} & \textbf{\#Samples}\\
    \hline
    ASR & OUAAL & Average WER & 47K \\
    \hline
    SQA & SD-QA-Syn & LLM as Judge & 1.2K \\
    \hline
    % AST & CoVoST-v2 (Ar2En) & Bleu  & 2.3K \\
    \multirow{2}{*}{AST} & \multirow{2}{*}{CoVoST-v2} & \multirow{2}{*}{\shortstack{Blue\\Sentence-Similarity}} & \multirow{2}{*}{2.3K} \\ & & & \\
    \hline
    DI & ADI-17 & Accuracy & 12K \\
    \hline
    SER & Synthesized Speech & Accuracy  & 1.1K \\
    \hline
    % AR & Common Voice+Elevenlabs-Syn & Accuracy  & 1.2K \\
    \multirow{2}{*}{AR} &  \multirow{2}{*}{\shortstack{Common Voice\\Elevenlabs-Syn}} & \multirow{2}{*}{Accuracy} & \multirow{2}{*}{1.2K} \\ & & & \\
    \hline
    GR & Common Voice & Accuracy  & 2K \\
    \hline
    \end{tabular}
}
\end{table}

% \onecolumn
\begin{table*}[htbp]
  \caption{Comprehensive performance benchmark of Nuha-Speech fine-tuned models and their original baselines.}
  \label{tab:results}
  \centering
  \resizebox{\textwidth}{!}{
  \begin{tabular}{c c c c c c c c c c}
    \hline
    & \multicolumn{2}{c}{\textbf{ASR}} & \multicolumn{2}{c}{\textbf{AST}} & \textbf{SQA} & \textbf{DI} & \textbf{SER} & \textbf{AR} & \textbf{GR} \\

    \textbf{Model} & WER\%$\downarrow$ & CER\%$\downarrow$ & Bleu$\uparrow$ & Sim$\uparrow$ & LLM-Score$\uparrow$ & Acc\%$\uparrow$ & Acc\%$\uparrow$ & Acc\%$\uparrow$ & Acc\%$\uparrow$ \\
    \hline
    Qwen2.5-omni-3B & - & - & 39.27 & 0.813 & 2.30 & 9.70 & 51.93 & 41.75 & 60.55 \\
    Nuha-Speech-3B & 36.83\% & 17.18 & 48.32 & 0.884 & 2.60 & 57.83 & \textbf{88.14} & 83.00 & 98.85 \\
    \hline
    Qwen2.5-omni-7B & 74.74 & 48.33 & 42.08 & 0.836 & 2.37 & 13.66 & 72.14 & 33.33 & 86.30 \\
    Nuha-Speech-7B & 34.91 & 16.09 & 50.60 & 0.896 & \textbf{2.66} & 72.65 & 86.64 & 73.00 & \textbf{99.50} \\
    \hline
    Qwen3-omni-30B & 30.71 & \textbf{13.67} & 48.71 & 0.889 & 2.59 & 16.78 & 80.23 & 42.67 & 82.25 \\
    Nuha-Speech-30B & \textbf{29.66} & 14.55 & \textbf{51.73} & \textbf{0.908} & 2.62 & \textbf{76.02} & 84.88 & \textbf{85.75} & 94.30\\

    \hline
  \end{tabular}
}
\end{table*}

Similar to the training set construction, the evaluation data also combines existing corpora with carefully curated samples.
For ASR evaluation, we followed the previously mentioned Open Universal Arabic ASR Leaderboard \cite{leaderboard}, a multi-dialect benchmark built on 6 datasets with 46,757 test samples. We report results in terms of average Word Error Rate (WER).
For AST, we use the Ar-En subset of the CoVoST-2 test set containing 2.3K samples. Evaluation metrics include BLEU\footnote{\url{https://pypi.org/project/sacrebleu/}} for lexical overlap and Gemma-based sentence similarity\footnote{\url{https://huggingface.co/google/embeddinggemma-300m}} to capture semantic alignment.
For the SQA task, we utilize the SD-QA \cite{sd_qa} Arabic textual QA dataset and synthesize its contextual passages into speech using the XTTS-v2 \cite{xtts}\footnote{\url{https://huggingface.co/coqui/XTTS-v2}} model with 58 distinct speaker voices. We then used Whisper-Large-v3\footnote{\url{https://huggingface.co/openai/whisper-large-v3}} to transcribe the synthesized speech and applied a CER-based filtering criterion to remove degraded speech synthesis outputs, yielding 1,237 high-quality synthesized utterances. As for evaluation, we employed LLaMA-3.3-70B-Instruct\footnote{\url{https://huggingface.co/meta-llama/Llama-3.3-70B-Instruct}} as an LLM judge, rating models' outputs in terms of relevance, correctness, and conciseness against the corresponding reference answers. This results in an overall score ranging from 0 to 3 for each model output.

For the dialect identification (DI) task, we directly incorporated the ADI-17\cite{ADI17} test set, which includes 12,150 samples evenly distributed over 17 dialect classes.
For SER, we followed the same data synthesis strategy as in the training set and generated another 1.1K emotional samples with 4 balanced categories.
For AR, we also sampled from the Common Voice test set and generated additional utterances using ElevenLabs. The final dataset comprises 1.2K samples evenly distributed across 3 age groups.
For GR, we selected 2,000 gender-balanced samples from the Common Voice test set.

For the four speech paralinguistic tasks described above, we uniformly adopted accuracy as the evaluation metric, any response falling outside the defined domain or predicting an excluded class is considered a misclassification.
The full evaluation scripts, LLM-as-Judge prompts and test sets used in our experiments will be made publicly available through GitHub\footnote{\url{https://github.com/Natural-Language-Processing-Elm/Nuha_Speech_Benchmark}}.

\subsection{Results}
Table 4 summarizes the performance of all evaluated models, including both the baseline and fine-tuned versions, across all seven tasks. From the model perspective, all models demonstrated improvements across all tasks after fine-tuning compared to the baseline. This improvement was more significant for the 3B and 7B models, whereas the 30B model showed smaller gains, since it has already obtained strong pre-trained semantic capabilities during pre-training. From the task perspective, paralinguistic tasks exhibited more substantial performance gains than semantic tasks after fine-tuning, suggesting a potential under-representation of Arabic speech paralinguistics tasks in the pre-training phase.

More specifically, for the ASR task, the Nuha-Speech-30B model maintained the solid performance that had already been established by its baseline. The Nuha-Speech-3B and Nuha-Speech-7B models both achieved significantly improved ASR capability in comparison with their baselines. For the AST task, all baseline models demonstrate a fundamental capability with reasonable BLEU scores and sentence similarities, and the Nuha-Speech-30B model consistently achieved the best performance across both metrics. Similarly, in the SQA evaluation, all baseline models attained an average score greater than two, with the fine-tuned Nuha-Speech-7B model marginally outperforming the other two. From the three semantic tasks above, we observe that although not all baseline models were directly trained on the Arabic ASR task, every model nonetheless acquired a certain degree of Arabic semantic understanding capacity, owing to the diversity of the training tasks through which Arabic semantics were introduced.

For speech paralinguistic tasks, we observe that baseline models initially exhibit very poor capability in dialect identification and age recognition, with accuracies close to random guessing. After fine-tuning, each model exhibited a dramatic, multi-fold increase in performance on both tasks, with Nuha-Speech-30B remaining the top-performing model. For the SER and GR tasks, the baseline models showed adequate initial performance, and after fine-tuning, they all further improved and reached excellent levels. 

The results obtained from both speech understanding and speech paralinguistics benchmarks demonstrate the effectiveness of our dataset design and confirm the validity of the adopted fine-tuning approach.

It should be noted that we also included Qwen2-audio-7B-instruct\footnote{\url{https://huggingface.co/Qwen/Qwen2-Audio-7B-Instruct}} as an additional baseline model in our experiments due to its broader set of speech tasks involved in pretraining. However, after being fine-tuned on the same training corpus, it consistently underperformed across all evaluated tasks compared to fine-tuning the Qwen2.5-omni-7B model which is of the same size. We therefore excluded it from our final benchmark to maintain a more competitive set of baseline models. By introducing the first systematic Arabic speech multi-task benchmark, we hope this benchmark will establish a standard for the evaluation of future Arabic Speech-LLMs.

\section{Conclusions}
In this paper, we present a collection of general-purpose Arabic speech-LLMs that, for the first time, provides unified support for broad Arabic speech tasks while enabling Arabic instruction-following. To achieve this goal, we constructed an Arabic speech-LLM training dataset under limited-resource conditions, while promoting reproducibility through the use of largely open-access datasets and transparent curation strategies. 
Additionally, we fully documented our fine-tuning protocols and established a common Arabic speech-LLM benchmark that jointly evaluates speech understanding and speech paralinguistics capabilities. 

Our work aims to lay the foundation for the Arabic speech-LLM infrastructure and to provide useful guidance for future research and development in building more capable general-purpose Arabic speech-LLMs.

\bibliography{custom}

% \appendix

% \section{Example Appendix}
% \label{sec:appendix}

% This is an appendix.

\end{document}